%% file: extreme-sum.tex
\documentclass[11pt,a4paper]{article}
\usepackage[hyperref]{emnlp2018}
\usepackage{times}
\usepackage{latexsym}

\usepackage{url}

\aclfinalcopy 
\def\aclpaperid{1090} 

\usepackage{amsmath}
\usepackage{mathtools}
\mathtoolsset{showonlyrefs}
\usepackage{amssymb}
\usepackage{multirow}

\usepackage{tree-dvips}
\usepackage{pgf,tikz}
\usepackage{tikz-qtree}
\usepackage{graphicx}

\usepackage{lipsum}  
\definecolor{forestgreen}{HTML}{009B55}
\definecolor{sepia}{HTML}{671800}
\definecolor{midnightblue}{HTML}{006795}
\definecolor{orangered}{HTML}{ED135A}

\title{Don't Give Me the Details, Just the Summary! \\Topic-Aware
  Convolutional Neural Networks for Extreme Summarization}

\author{Shashi Narayan \quad Shay B. Cohen \quad Mirella Lapata \\ 
  Institute for Language, Cognition and Computation \\
  School of Informatics, University of Edinburgh \\ 
  10 Crichton Street, Edinburgh, EH8 9AB  \\ 
  \texttt{\url{shashi.narayan@ed.ac.uk}},
  \texttt{\{scohen,mlap\}@inf.ed.ac.uk} 
}

\makeatletter
\newcommand{\thickhline}{%
    \noalign {\ifnum 0=`}\fi \hrule height 1pt
    \futurelet \reserved@a \@xhline
}
\makeatother

\date{}

\begin{document}
\maketitle
\begin{abstract}

  We introduce \emph{extreme summarization}, a new single-document
  summarization task which does not favor extractive strategies and
  calls for an abstractive modeling approach. The idea is to create a
  short, one-sentence news summary answering the question ``What is
  the article about?''. We collect a real-world, large scale dataset
  for this task by harvesting online articles from the British
  Broadcasting Corporation (BBC). We propose a novel abstractive model
  which is conditioned on the article's topics and based entirely on
  convolutional neural networks.  We demonstrate experimentally that
  this architecture captures long-range dependencies in a document and
  recognizes pertinent content, outperforming an oracle extractive
  system and state-of-the-art abstractive approaches when evaluated
  automatically and by humans.\footnote{Our dataset, code, and demo
    are available at: \url{https://github.com/shashiongithub/XSum}.}

\end{abstract}

\input{introduction}

\input{xsum-dataset}

\input{model}

\input{setup-evaluations}

\section{Conclusions}

In this paper we introduced the task of ``extreme summarization''
together with a large-scale dataset which pushes the boundaries of
abstractive methods.  Experimental evaluation revealed that models
which have abstractive capabilities do better on this task and that
high-level document knowledge in terms of topics and long-range
dependencies is critical for recognizing pertinent content and
generating informative summaries. In the future, we would like to
create more linguistically-aware encoders and decoders incorporating
co-reference and entity linking.

\vspace{-0.1cm}

\paragraph{Acknowledgments} We gratefully acknowledge the support of
the European Research Council (Lapata; award number 681760), the
European Union under the Horizon 2020 SUMMA project (Narayan, Cohen;
grant agreement 688139), and Huawei Technologies (Cohen).

\bibliography{summarisation-improved}
\bibliographystyle{acl_natbib_nourl}

\end{document}

%% file: introduction.tex
\section{Introduction}

Automatic summarization is one of the central problems in Natural
Language Processing (NLP) posing several challenges relating to {\em
  understanding} (i.e.,~identifying important content) and {\em
  generation} (i.e.,~aggregating and rewording the identified
content into a summary). Of the many summarization paradigms that have
been identified over the years (see \citeauthor{mani2001automatic},
\citeyear{mani2001automatic} and \citeauthor{Nenkova:McKeown:2011},
\citeyear{Nenkova:McKeown:2011} for a comprehensive overview),
single-document summarization has consistently attracted attention
\cite{jp-acl16,durrett-nyt-ext,nallapati-signll16,nallapati17,see-acl17,tanwan-acl17,narayan-arxiv17,Fan2017Controllable,paulus-socher-arxiv17,Pasunuru-multireward18,asli-multiagent18,narayan-sidenet18,narayan-rank18}.

\begin{figure}[t!]
  \center{\fontsize{10}{12}\selectfont 
    \begin{tabular}{ p{7.2cm} }
      \thickhline
      
      
      \textbf{\textsc{Summary:}} \emph{\textcolor{orangered}{A man and
          a child have been killed} after a \textcolor{sepia}{light}
        \textcolor{forestgreen}{aircraft made an emergency landing} on
        \textcolor{midnightblue}{a beach in Portugal}.}  
       \\ \thickhline
      
      \textbf{\textsc{Document:}} Authorities said the incident took place on \textcolor{midnightblue}{Sao Joao beach in Caparica, south-west of Lisbon}. \\ 

      The National Maritime Authority said \textcolor{orangered}{a middle-aged man and a young girl died} after they were unable to avoid the plane. \\
      
      [{\em 6 sentences with 139 words are abbreviated from here.}] \\
      




      Other reports said the victims had been sunbathing when \textcolor{forestgreen}{the plane made its emergency landing}. \\

      [{\em Another 4 sentences with 67 words are abbreviated from here.}] \\



      Video footage from the scene carried by local broadcasters showed \textcolor{sepia}{a small recreational plane} parked on the sand, apparently intact and surrounded by beachgoers and emergency workers. 
      \\

        
      [{\em Last 2 sentences with 19 words are abbreviated.}] \\
      
      \thickhline
    \end{tabular}     
  }
  \caption{An abridged example from our extreme summarization dataset
    showing the document and its one-line summary. Document content
    present in the summary is color-coded.
  }\label{fig:bbcex-1}
\end{figure}

Neural approaches to NLP and their ability to learn continuous
features without recourse to pre-processing tools or linguistic
annotations have driven the development of large-scale document
summarization datasets
\cite{nytcorpus,hermann-nips15,newsroom-naacl18}.
However, these datasets often favor extractive models which create a
summary by identifying (and subsequently concatenating) the most
important sentences in a document
\cite{jp-acl16,nallapati17,narayan-rank18}.  Abstractive approaches,
despite being more faithful to the actual summarization task, either
lag behind extractive ones or are mostly extractive, exhibiting a
small degree of abstraction
\cite{see-acl17,tanwan-acl17,paulus-socher-arxiv17,Pasunuru-multireward18,asli-multiagent18}.

In this paper we introduce \emph{extreme summarization}, a new
single-document summarization task which is not amenable to extractive
strategies and requires an abstractive modeling approach. The idea is
to create a short, one-sentence news summary answering the question
``What is the article about?''. An example of a document and its
extreme summary are shown in Figure~\ref{fig:bbcex-1}. As can be seen,
the summary is very different from a headline whose aim is to
encourage readers to read the story; it draws on information
interspersed in various parts of the document (not only the beginning)
and displays multiple levels of abstraction including paraphrasing,
fusion, synthesis, and inference. We build a dataset for the proposed
task by harvesting online articles from the British Broadcasting
Corporation (BBC) that often include a first-sentence summary.



We further propose a novel deep learning model which we argue is
well-suited to the extreme summarization task. Unlike most existing
abstractive approaches
\cite{rush-acl15,chenIjcai-16,nallapati-signll16,see-acl17,tanwan-acl17,paulus-socher-arxiv17,Pasunuru-multireward18,asli-multiagent18}
which rely on an encoder-decoder architecture modeled by recurrent
neural networks (RNNs), we present a \emph{topic-conditioned} neural
model which is based entirely on convolutional neural networks
\cite{convseq2seq}. Convolution layers capture long-range
dependencies between words in the document more effectively compared to
RNNs, allowing to perform document-level inference, abstraction, and
paraphrasing.  Our convolutional encoder associates each word with a
topic vector capturing whether it is representative of the document's
content, while our convolutional decoder conditions each word
prediction on a document topic vector.

%


Experimental results show that when evaluated automatically (in terms
of ROUGE) our topic-aware convolutional model outperforms an oracle
extractive system and state-of-the-art RNN-based abstractive
systems. We also conduct two human evaluations in order to assess (a)
which type of summary participants prefer and (b) how much key
information from the document is preserved in the summary. Both
evaluations overwhelmingly show that human subjects find our summaries
more informative and complete. Our contributions in this work are
three-fold: a new single document summarization dataset that
encourages the development of abstractive systems; corroborated by
analysis and empirical results showing that extractive approaches are
not well-suited to the extreme summarization task; and a novel
topic-aware convolutional sequence-to-sequence model for abstractive
summarization.


%% file: xsum-dataset.tex
\section{The XSum Dataset}
\label{sec:bbcdataset}

\begin{table*}
  \begin{center}{\footnotesize
  \begin{tabular}{ l | l | c c | c c | r r}
    \thickhline
    \multirow{2}{*}{Datasets} & \multirow{2}{*}{\# docs (train/val/test)} & \multicolumn{2}{c|}{avg. document length} & \multicolumn{2}{c|}{avg. summary length} & \multicolumn{2}{c}{vocabulary size}\\
    & & words & sentences & words & sentences & document & summary\\ \thickhline 
    CNN & 90,266/1,220/1,093 & 760.50 & 33.98 & 45.70 & 3.59 & 343,516 & 89,051 \\
    DailyMail & 196,961/12,148/10,397 & 653.33 & 29.33 & 54.65 & 3.86 & 563,663 & 179,966 \\
    NY Times & 589,284/32,736/32,739 & 800.04 & 35.55 & 45.54 & 2.44 & 1,399,358 & 294,011 \\
    XSum & 204,045/11,332/11,334 & 431.07 & 19.77 & 23.26 & 1.00 & 399,147 & 81,092 \\ \thickhline
  \end{tabular}}
  \end{center}
  \caption{Comparison of summarization datasets
    with respect to overall corpus size, size of training, validation,
    and test set, average document (source) and summary (target)
    length (in terms of words and sentences), and  vocabulary size on
    both on source and target. For CNN and DailyMail,
    we  used the original splits of
    \protect\newcite{hermann-nips15} and followed
    \protect\newcite{narayan-rank18} to preprocess them. For  NY
    Times  \protect\cite{nytcorpus}, we  used the splits
    and pre-processing steps of
    \protect\newcite{paulus-socher-arxiv17}. For the vocabulary, we
    lowercase tokens.} \label{table:bbc-size-comparison} 
\end{table*}

\begin{table*}
  \begin{center}{\footnotesize
  \begin{tabular}{ l | c c c c | c c c | c c c } 
    \thickhline
    \multirow{2}{*}{Datasets} & \multicolumn{4}{c|}{\% of novel n-grams in gold summary} & \multicolumn{3}{c|}{\textsc{lead}} & \multicolumn{3}{c}{\textsc{ext-oracle}} \\
    & unigrams & bigrams & trigrams & 4-grams & R1 & R2 & RL & R1 & R2 & RL \\ \thickhline 
    CNN & 16.75 & 54.33 & 72.42 & 80.37 & 29.15 & 11.13 & 25.95 & 50.38 & 28.55 & 46.58 \\
    DailyMail & 17.03 & 53.78 & 72.14 & 80.28 & 40.68 & 18.36 & 37.25 & 55.12 & 30.55 & 51.24 \\
    NY Times & 22.64 & 55.59 & 71.93 & 80.16 & 31.85 & 15.86 & 23.75 & 52.08 & 31.59 & 46.72 \\
    XSum & \textbf{35.76} & \textbf{83.45}  & \textbf{95.50}  & \textbf{98.49}  & \textbf{16.30} & \textbf{1.61}  & \textbf{11.95}  & \textbf{29.79}  & \textbf{8.81}  & \textbf{22.65}  \\ \thickhline
  \end{tabular}}
  \end{center}
  \caption{Corpus bias towards extractive methods in the CNN,
    DailyMail, NY Times, and XSum datasets.  We show  the proportion of novel $n$-grams in gold summaries. We also report ROUGE scores  for
    the  \textsc{lead} baseline and the extractive oracle system
    \textsc{ext-oracle}.     Results
    are computed on the test set.
    \label{table:ngram-coverage-lead-oracle}}
\end{table*}


Our extreme summarization dataset (which we call XSum) consists of BBC
articles and accompanying single sentence summaries. Specifically,
each article is prefaced with an introductory sentence (aka summary)
which is professionally written, typically by the author of the
article. The summary bears the HTML class
``story-body\_\_introduction,'' and can be easily identified and
extracted from the main text body (see Figure~\ref{fig:bbcex-1} for an
example summary-article pair).



We followed the methodology proposed in \newcite{hermann-nips15} to
create a large-scale dataset for extreme summarization. Specifically,
we collected~226,711 Wayback archived BBC articles ranging over almost
a decade (2010 to 2017) and covering a wide variety of domains
(e.g.,~News, Politics, Sports, Weather, Business, Technology, Science,
Health, Family, Education, Entertainment and Arts). Each article comes
with a unique identifier in its URL, which we used to randomly split
the dataset into training (90\%, 204,045), validation (5\%, 11,332),
and test (5\%, 11,334) set. Table~\ref{table:bbc-size-comparison}
compares XSum with the CNN, DailyMail, and NY Times 
benchmarks. As can be seen, XSum contains a substantial number of
training instances, similar to DailyMail; documents and summaries in
XSum are shorter in relation to other datasets but the vocabulary size
is sufficiently large, comparable to CNN.


Table~\ref{table:ngram-coverage-lead-oracle} provides empirical
analysis supporting our claim that XSum is less biased toward
extractive methods compared to other summarization datasets.  We
report the percentage of novel $n$-grams in the target gold summaries
that do not appear in their source documents.  There are 36\% novel
unigrams in the XSum reference summaries compared to 17\% in CNN,
17\%~in DailyMail, and 23\%~in NY Times. 
This indicates that XSum summaries are more abstractive. The
proportion of novel constructions grows for larger $n$-grams across
datasets, however, it is much steeper in XSum whose summaries exhibit
approximately 83\% novel bigrams, 96\% novel trigrams, and 98\% novel
4-grams (comparison datasets display around 47--55\%~new bigrams,
58--72\%~new trigrams, and 63--80\%~novel 4-grams).

We further evaluated two extractive methods on these datasets.
\textsc{lead} is often used as a strong lower bound for news
summarization \cite{nenkova-05} and creates a summary by selecting the
first few sentences or words in the document. We extracted the first
3~sentences for CNN documents and the first 4~sentences for DailyMail
\cite{narayan-rank18}. Following previous work
\cite{durrett-nyt-ext,paulus-socher-arxiv17}, we obtained
\textsc{lead} summaries based on the first 100 words for NY Times
documents. 
For XSum, we selected the first sentence
in the document (excluding the one-line summary) to generate the
\textsc{lead}. Our second method, \textsc{ext-oracle}, can be viewed
as an upper bound for extractive models
\cite{nallapati17,narayan-rank18}. It creates an oracle summary by
selecting the best possible set of sentences in the document that
gives the highest ROUGE \cite{rouge} with respect to the gold
summary. For XSum, we simply selected the single-best sentence in the
document as summary.

Table~\ref{table:ngram-coverage-lead-oracle} reports the performance
of the two extractive methods using ROUGE-1 (R1), ROUGE-2 (R2), and
ROUGE-L (RL) with the gold summaries as reference. The \textsc{lead}
baseline performs extremely well on CNN, DailyMail and NY
Times 
confirming that they are biased towards extractive methods.
\textsc{ext-oracle} further shows that improved sentence selection
would bring further performance gains to extractive
approaches. Abstractive systems trained on these datasets often have a
hard time beating the \textsc{lead}, let alone \textsc{ext-oracle}, or
display a low degree of novelty in their summaries
\cite{see-acl17,tanwan-acl17,paulus-socher-arxiv17,Pasunuru-multireward18,asli-multiagent18}.
Interestingly, \textsc{lead} and \textsc{ext-oracle} perform poorly on
XSum underlying the fact that it is less biased towards extractive
methods.


In line with our findings, \newcite{newsroom-naacl18} have recently
reported similar extractive biases in existing datasets. They
constructed a new dataset called ``Newsroom'' which demonstrates a
high diversity of summarization styles. XSum is not diverse, it
focuses on a single news outlet (i.e.,~BBC) and a unifrom
summarization style (i.e.,~a single sentence). However, it is
sufficiently large for neural network training and we hope it will
spur further research towards the development of abstractive
summarization models.




%% file: model.tex
\section{Convolutional Sequence-to-Sequence Learning for Summarization}
\label{sec:topicconvabssum}

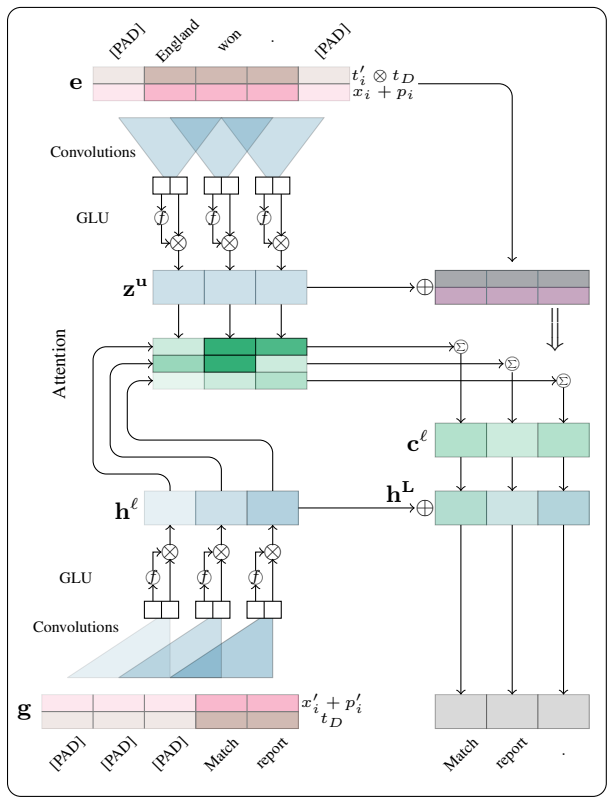
\begin{figure}[t!]
  \center{\footnotesize
    \begin{tikzpicture}[scale=0.45]
      
      \begin{scope}[shift={(0,0)}] 
        \draw [rounded corners=5pt] (-2.5,1.75) rectangle (15,-21.5);
        
        \draw [fill=sepia,opacity=0.1] (0,0) rectangle (1.5,-0.5);
        \draw [fill=sepia,opacity=0.3] (1.5,0) rectangle (6,-0.5);
        \draw [fill=sepia,opacity=0.1] (6,0) rectangle (7.5,-0.5);
        \draw [fill=orangered,opacity=0.1] (0,-0.5) rectangle (1.5,-1);
        \draw [fill=orangered,opacity=0.3] (1.5,-0.5) rectangle (6,-1);
        \draw [fill=orangered,opacity=0.1] (6,-0.5) rectangle (7.5,-1);
        
        \draw [gray] (0,-0.5) -- (7.5,-0.5);
        \draw [gray] (1.5,0) -- (1.5,-1);
        \draw [gray] (3,0) -- (3,-1);
        \draw [gray] (4.5,0) -- (4.5,-1);
        \draw [gray] (6,0) -- (6,-1);
        \node at (8.5,-0.75) {\tiny $x_i+p_i$};
        \node at (8.5,-0.25) {\tiny $t'_i \otimes t_D$};  
        \node at (-0.5,-0.5) {$\mathbf{e}$};

        \node at (1,0.75) {\tiny \rotatebox{45}{[PAD]}}; 
        \node at (2.5,0.75) {\tiny \rotatebox{45}{England}}; 
        \node at (4,0.75) {\tiny \rotatebox{45}{won}}; 
        \node at (5.25,0.75) {\tiny \rotatebox{45}{.}}; 
        \node at (7,0.75) {\tiny \rotatebox{45}{[PAD]}}; 
        
        
        \node at (0,-2.5) {\tiny Convolutions};
        
        \draw [fill=midnightblue,opacity=0.2] (3.75,-1.5) -- (6.75, -1.5) -- (5.25, -3.5) -- (3.75,-1.5);
        \draw [fill=midnightblue,opacity=0.2] (2.25,-1.5) -- (5.25, -1.5) -- (3.75, -3.5) -- (2.25,-1.5);
        \draw [fill=midnightblue,opacity=0.2] (0.75,-1.5) -- (3.75, -1.5) -- (2.25, -3.5) -- (0.75,-1.5);
        
        \draw [fill=white,opacity=1] (1.75,-3.25) rectangle (2.75,-3.75);
        \draw [fill=white,opacity=1] (3.25,-3.25) rectangle (4.25,-3.75);
        \draw [fill=white,opacity=1] (4.75,-3.25) rectangle (5.75,-3.75);
        
        \draw (2.25,-3.25) -- (2.25,-3.75);
        \draw (3.75,-3.25) -- (3.75,-3.75);
        \draw (5.25,-3.25) -- (5.25,-3.75);
        

        \node at (0,-4.45) {\tiny GLU};
        
        \draw [->] (2,-3.75) -- (2,-4.25);
        \draw [gray] (2,-4.45) circle (0.2cm);
        \node at (2, -4.45) {\tiny $f$};
        \draw [->] (2.5,-3.75) -- (2.5,-5);
        \node at (2.5, -5.2) {$\otimes$};
        \draw [->] (2, -4.65) -- (2, -5.2) -- (2.25,-5.2);
        \draw [->] (2.5,-5.4) -- (2.5,-6);

        \draw [->] (3.5,-3.75) -- (3.5,-4.25);
        \draw [gray] (3.5, -4.45) circle (0.2cm);
        \node at (3.5, -4.45) {\tiny $f$};
        \draw [->] (4,-3.75) -- (4,-5);
        \node at (4,-5.2) {$\otimes$};
        \draw [->] (3.5,-4.65) -- (3.5,-5.2) -- (3.75,-5.2);
        \draw [->] (4,-5.4) -- (4,-6);

        \draw [->] (5,-3.75) -- (5,-4.25);
        \draw [gray] (5, -4.45) circle (0.2cm);
        \node at (5, -4.45) {\tiny $f$};
        \draw [->] (5.5,-3.75) -- (5.5,-5);
        \node at (5.5, -5.2) {$\otimes$};
        \draw [->] (5, -4.65) -- (5,-5.2) -- (5.25,-5.2);
        \draw [->] (5.5,-5.4) -- (5.5,-6);

        \draw [fill=midnightblue,opacity=0.2] (1.75,-6) rectangle (3.25,-7);
        \draw [fill=midnightblue,opacity=0.2] (3.25,-6) rectangle (4.75,-7);
        \draw [fill=midnightblue,opacity=0.2] (4.75,-6) rectangle (6.25,-7);
        \node at (1.25,-6.5) {$\mathbf{z^u}$};

        \node at (-1,-8.75) {\scriptsize \rotatebox{90}{Attention}}; 

        \draw [->] (2.5,-7) -- (2.5,-8);
        \draw [->] (4,-7) -- (4,-8);
        \draw [->] (5.5,-7) -- (5.5,-8);

        \draw [fill=forestgreen,opacity=0.2] (1.75,-8) rectangle (3.25,-8.5);
        \draw [fill=forestgreen,opacity=0.8] (3.25,-8) rectangle (4.75,-8.5);
        \draw [fill=forestgreen,opacity=0.7] (4.75,-8) rectangle (6.25,-8.5);
        \draw [fill=forestgreen,opacity=0.4] (1.75,-8.5) rectangle (3.25,-9);
        \draw [fill=forestgreen,opacity=0.8] (3.25,-8.5) rectangle (4.75,-9);
        \draw [fill=forestgreen,opacity=0.2] (4.75,-8.5) rectangle (6.25,-9);
        \draw [fill=forestgreen,opacity=0.1] (1.75,-9) rectangle (3.25,-9.5);
        \draw [fill=forestgreen,opacity=0.2] (3.25,-9) rectangle (4.75,-9.5);
        \draw [fill=forestgreen,opacity=0.3] (4.75,-9) rectangle (6.25,-9.5);

        \draw [fill=sepia,opacity=0.3] (10,-6) rectangle (14.5,-6.5);
        \draw [fill=orangered,opacity=0.3] (10,-6.5) rectangle (14.5,-7);
        \draw [gray] (11.5,-6) -- (11.5,-7);
        \draw [gray] (13,-6) -- (13,-7);
        \draw [fill=midnightblue,opacity=0.2] (10,-6) rectangle (14.5,-7);
        
        \draw [rounded corners=5pt,->] (9.5,-0.5) -- (12.25, -0.5) -- (12.25, -5.75);
        \draw [->] (6.25,-6.5) -- (9.4, -6.5);
        \node at (9.7,-6.5) {$\oplus$};
      \end{scope}

      \begin{scope}[shift={(-1.5,-19.5)}] 
        
        \draw [fill=sepia,opacity=0.1] (0,0) rectangle (4.5,0.5);
        \draw [fill=sepia,opacity=0.3] (4.5,0) rectangle (7.5,0.5);        
        \draw [fill=orangered,opacity=0.1] (0,0.5) rectangle (4.5,1);
        \draw [fill=orangered,opacity=0.3] (4.5,0.5) rectangle (7.5,1);

        \node at (0.75,-0.75) {\tiny \rotatebox{45}{[PAD]}}; 
        \node at (2.25,-0.75) {\tiny \rotatebox{45}{[PAD]}}; 
        \node at (3.75,-0.75) {\tiny \rotatebox{45}{[PAD]}}; 
        \node at (5.25,-0.75) {\tiny \rotatebox{45}{Match}}; 
        \node at (6.75,-0.75) {\tiny \rotatebox{45}{report}};

        \draw [gray] (0,0.5) -- (7.5,0.5);
        \draw [gray] (1.5,0) -- (1.5,1);
        \draw [gray] (3,0) -- (3,1);
        \draw [gray] (4.5,0) -- (4.5,1);
        \draw [gray] (6,0) -- (6,1);
        \node at (8.5,0.75) {\tiny $x'_i+p'_i$};
        \node at (8.5,0.25) {\tiny $t_D$};        
        \node at (-0.5,0.5) {$\mathbf{g}$};

        
        \node at (1,3) {\tiny Convolutions};

        \draw [fill=midnightblue,opacity=0.3] (3.75,1.5) -- (6.75, 1.5) -- (6.75, 3.5) -- (3.75,1.5);
        \draw [fill=midnightblue,opacity=0.2] (2.25,1.5) -- (5.25, 1.5) -- (5.25, 3.5) -- (2.25,1.5);
        \draw [fill=midnightblue,opacity=0.1] (0.75,1.5) -- (3.75, 1.5) -- (3.75, 3.5) -- (0.75,1.5);
        
        \draw [fill=white,opacity=1] (3,3.25) rectangle (4,3.75);
        \draw [fill=white,opacity=1] (4.5,3.25) rectangle (5.5,3.75);
        \draw [fill=white,opacity=1] (6,3.25) rectangle (7,3.75);
        
        \draw (3.5,3.25) -- (3.5,3.75);
        \draw (5,3.25) -- (5,3.75);
        \draw (6.5,3.25) -- (6.5,3.75);
        
        \node at (1,4.45) {\tiny GLU};

        \draw [->] (3.25,3.75) -- (3.25,4.25);
        \draw [gray] (3.25, 4.45) circle (0.2cm);
        \node at (3.25, 4.45) {\tiny $f$};
        \draw [->] (3.75,3.75) -- (3.75,5);
        \node at (3.75, 5.2) {$\otimes$};
        \draw [->] (3.25, 4.65) -- (3.25, 5.2) -- (3.5,5.2);
        \draw [->] (3.75,5.4) -- (3.75,6);

        \draw [->] (4.75,3.75) -- (4.75,4.25);
        \draw [gray] (4.75, 4.45) circle (0.2cm);
        \node at (4.75, 4.45) {\tiny $f$};
        \draw [->] (5.25,3.75) -- (5.25,5);
        \node at (5.25, 5.2) {$\otimes$};
        \draw [->] (4.75, 4.65) -- (4.75,5.2) -- (5,5.2);
        \draw [->] (5.25,5.4) -- (5.25,6);

        \draw [->] (6.25,3.75) -- (6.25,4.25);
        \draw [gray] (6.25, 4.45) circle (0.2cm);
        \node at (6.25, 4.45) {\tiny $f$};
        \draw [->] (6.75,3.75) -- (6.75,5);
        \node at (6.75, 5.2) {$\otimes$};
        \draw [->] (6.25, 4.65) -- (6.25,5.2) -- (6.5,5.2);
        \draw [->] (6.75,5.4) -- (6.75,6);

        \draw [fill=midnightblue,opacity=0.1] (3,6) rectangle (4.5,7);
        \draw [fill=midnightblue,opacity=0.2] (4.5,6) rectangle (6,7);
        \draw [fill=midnightblue,opacity=0.3] (6,6) rectangle (7.5,7);
        
        \node at (2.5,6.5) {$\mathbf{h^{\ell}}$};

        \draw [fill=forestgreen,opacity=0.3] (11.5,8) rectangle (13,9);
        \draw [fill=forestgreen,opacity=0.2] (13,8) rectangle (14.5,9);
        \draw [fill=forestgreen,opacity=0.3] (14.5,8) rectangle (16,9);     
        \draw [->] (12.25,8) -- (12.25,7);     
        \draw [->] (13.75,8) -- (13.75,7);     
        \draw [->] (15.25,8) -- (15.25,7); 
        \node at (11,8.5) {$\mathbf{c^{\ell}}$};

        \draw [fill=midnightblue!20!forestgreen,opacity=0.3] (11.5,6) rectangle (13,7);
        \draw [fill=midnightblue!60!forestgreen,opacity=0.2] (13,6) rectangle (14.5,7);
        \draw [fill=midnightblue!80!forestgreen,opacity=0.3] (14.5,6) rectangle (16,7);        
        \draw [->] (7.5,6.5) -- (10.9, 6.5);
        \node at (11.2,6.5) {$\oplus$}; 
        \node at (10.5,7) {$\mathbf{h^L}$};
        
        \draw [->] (12.25,6) -- (12.25,1); 
        \draw [fill=gray,opacity=0.3] (11.5,0) rectangle (13,1);
        \draw [->] (13.75,6) -- (13.75,1);     
        \draw [fill=gray,opacity=0.3] (13,0) rectangle (14.5,1);
        \draw [->] (15.25,6) -- (15.25,1);     
        \draw [fill=gray,opacity=0.3] (14.5,0) rectangle (16,1);
        
        \node at (12.25,-0.75) {\tiny \rotatebox{45}{Match}}; 
        \node at (13.75,-0.75) {\tiny \rotatebox{45}{report}}; 
        \node at (15.25,-0.75) {\tiny \rotatebox{45}{.}};

        \draw [rounded corners=5pt,->] (3.75,7) -- (3.75,7.5) -- (1.5, 7.5) -- (1.5, 11.25) -- (3.25, 11.25);
        \draw [rounded corners=5pt,->] (5.25,7) -- (5.25,8) -- (2, 8) -- (2, 10.75) -- (3.25, 10.75);
        \draw [rounded corners=5pt,->] (6.75,7) -- (6.75,8.5) -- (2.5, 8.5) -- (2.5, 10.25) -- (3.25, 10.25);
        \draw [gray] (12.25,11.25) circle (0.2cm);
        \node at (12.25,11.25) {\fontsize{3}{5}\selectfont $\sum$};
        \draw [->] (7.75, 11.25) -- (12.05,11.25);
        \draw [->] (12.25, 11.05) -- (12.25,9);

        \draw [gray] (13.75,10.75) circle (0.2cm);
        \node at (13.75,10.75) {\fontsize{3}{5}\selectfont $\sum$};
        \draw [->] (7.75, 10.75) -- (13.5,10.75);
        \draw [->] (13.75, 10.5) -- (13.75,9);
        
        \draw [gray] (15.25,10.25) circle (0.2cm);
        \node at (15.25,10.25) {\fontsize{3}{5}\selectfont $\sum$};
        \draw [->] (7.75, 10.25) -- (15.05,10.25);
        \draw [->] (15.25, 10.05) -- (15.25,9);
        
        \node at (15,11.75) {$\Big\Downarrow$};
      \end{scope}
    \end{tikzpicture}
  }
  \vspace{-0.5cm}
  \caption{Topic-conditioned convolutional model for extreme
    summarization.}\label{fig:architecture}
\end{figure}

Unlike tasks like machine translation and paraphrase generation where
there is often a one-to-one semantic correspondence between source and
target words, document summarization must distill the content of the
document into a few important facts. This is even more challenging for
our task, where the compression ratio is extremely high, and pertinent
content can be easily missed.


Recently, a convolutional alternative to sequence modeling has been
proposed showing promise for machine translation
\cite{convenc_mt,convseq2seq} and story generation
\cite{fan-hier-gen}. We believe that convolutional architectures are
attractive for our summarization task for at least two
reasons. Firstly, contrary to recurrent networks which view the input
as a chain structure, convolutional networks can be stacked to
represent large context sizes. Secondly, hierarchical features can be
extracted over larger and larger contents, allowing to represent
long-range dependencies efficiently through shorter paths.

Our model builds on the work of \newcite{convseq2seq} who develop an
encoder-decoder architecture for machine translation with an attention
mechanism \cite{NIPS2015_5846} based exclusively on deep convolutional
networks. We adapt this model to our summarization task by allowing it
to recognize pertinent content (i.e.,~by foregrounding salient words
in the document).  In particular, we improve the convolutional encoder
by associating each word with a vector representing topic salience,
and the convolutional decoder by conditioning each word prediction on
the document topic vector.


\paragraph{Model Overview}

At the core of our model is a simple convolutional block structure
that computes intermediate states based on a fixed number of input
elements. Our convolutional encoder (shown at the top of Figure
\ref{fig:architecture}) applies this unit across the document. We
repeat these operations in a stacked fashion to get a multi-layer
hierarchical representation over the input document where words at
closer distances interact at lower layers while distant words interact
at higher layers. The interaction between words through hierarchical
layers effectively captures long-range dependencies. 

Analogously, our convolutional decoder (shown at the bottom of
Figure~\ref{fig:architecture}) uses the multi-layer convolutional
structure to build a hierarchical representation over what has been
predicted so far. Each layer on the decoder side determines useful
source context by attending to the encoder representation before it
passes its output to the next layer. This way the model remembers
which words it previously attended to and applies multi-hop attention
(shown at the middle of Figure \ref{fig:architecture}) per time
step. The output of the top layer is passed to a softmax classifier to
predict a distribution over the target vocabulary.

Our model assumes access to word and document topic
distributions. These can be obtained by any topic model, however we
use Latent Dirichlet Allocation (LDA; \citealt{Blei:2003:LDA}) in our
experiments; we pass the distributions obtained from LDA directly to
the network as additional input. This allows us to take advantage of
topic modeling without interfering with the computational advantages
of the convolutional architecture. The idea of capturing
document-level semantic information has been previously explored for
recurrent neural networks
\cite{mikolovZ12,Ghosh2016ContextualL,dieng-iclr17}, however, we are
not aware of any existing convolutional models.

\paragraph{Topic Sensitive Embeddings}

Let~$D$ denote a document consisting of a sequence of words $(w_1,
\ldots, w_m)$; we embed~$D$ into a distributional space $\mathbf{x} =
(x_1, \ldots, x_m)$ where $x_i \in \mathbb{R}^f$ is a column in
embedding matrix $M \in \mathbb{R}^{V\times f}$ (where $V$ is the
vocabulary size). We also embed the absolute word positions in the
document \mbox{$\mathbf{p} = (p_1, \ldots, p_m)$} where $p_i \in
\mathbb{R}^f$ is a column in position matrix $P \in
\mathbb{R}^{N\times f}$, and $N$ is the maximum number of
positions. Position embeddings have proved useful for convolutional
sequence modeling \cite{convseq2seq}, because, in contrast to RNNs,
they do not observe the temporal positions of words
\cite{D16-1248}. Let $t_D \in \mathbb{R}^{f'}$ be the topic
distribution of document~$D$ and $\mathbf{t'} = (t'_1, \ldots, t'_m)$
the topic distributions of words in the document (where $t'_i \in
\mathbb{R}^{f'}$). During encoding, we represent document~$D$ via
$\mathbf{e} = (e_1, \ldots, e_m)$, where~$e_i$ is:
\begin{align}
e_i = [(x_i+p_i);(t'_i\otimes t_D)] \in \mathbb{R}^{f+f'}, \label{eq:encoder}
\end{align}
\noindent and $\otimes$ denotes point-wise multiplication. The topic
distribution $t'_i$ of word $w_i$ essentially captures how topical the
word is in itself (local context), whereas the topic
distribution~$t_D$ represents the overall theme of the document
(global context). The encoder essentially enriches the context of the
word with its topical relevance to the document.

For every output prediction, the decoder estimates representation
$\mathbf{g} = (g_1, \ldots, g_n)$ for previously predicted words
$(w'_1, \ldots, w'_n)$ where $g_i$ is:
\begin{align}
g_i = [(x'_i+p'_i);t_D] \in \mathbb{R}^{f+f'}, \label{eq:decoder}
\end{align}
\noindent $x'_i$ and $p'_i$ are word and position embeddings of
previously predicted word~$w'_i$, and $t_D$ is the topic distribution
of the input document. Note that the decoder does not use the topic
distribution of~$w'_i$ as computing it on the fly would be
expensive. However, every word prediction is conditioned on the topic
of the document, enforcing the summary to have the same theme as the
document.


\paragraph{Multi-layer Convolutional Structure}

Each convolution block, parametrized by $W \in \mathbb{R}^{2d\times
  kd}$ and $b_w \in \mathbb{R}^{2d}$, takes as input $X \in
\mathbb{R}^{k\times d}$ which is the concatenation of $k$ adjacent
elements embedded in a~$d$ dimensional space, applies one dimensional
convolution and returns an output element $Y \in \mathbb{R}^{2d}$. We
apply Gated Linear Units (GLU, $v:\mathbb{R}^{2d} \rightarrow
\mathbb{R}^{d}$, \citeauthor{pmlr-v70-dauphin17a}
\citeyear{pmlr-v70-dauphin17a}) on the output of the convolution
$Y$. Subsequent layers operate over the $k$ output elements of the
previous layer and are connected through residual connections
\cite{He2016DeepRL} to allow for deeper hierarchical
representation. We denote the output of the $\ell$th layer as
$\mathbf{h^{\ell}} = (h^{\ell}_1, \ldots, h^{\ell}_n)$ for the decoder
network, and $\mathbf{z^{\ell}} = (z^{\ell}_1, \ldots, z^{\ell}_m)$
for the encoder network.

\paragraph{Multi-hop Attention}

Our encoder and decoder are tied to each other through a multi-hop
attention mechanism. For each decoder layer $\ell$, we compute the
attention $a^{\ell}_{ij}$ of state $i$ and source element $j$ as:
\begin{align}
a^{\ell}_{ij} = \frac{\mbox{exp}(d^{\ell}_i \cdot z^u_j)}{\sum^m_{t=1} \mbox{exp}(d^{\ell}_i \cdot z^u_t)}, \label{eq:attention}
\end{align}
\noindent where $d^{\ell}_i = W^{\ell}_dh^{\ell}_i+b^{\ell}_i+g_i$ is
the decoder state summary combining the current decoder state
$h^{\ell}_i$ and the previous output element embedding $g_i$. The
vector $\mathbf{z^u}$ is the output from the last encoder layer
$u$. The conditional input $c^{\ell}_i$ to the current decoder layer
is a weighted sum of the encoder outputs as well as the input element
embeddings $e_j$:
\begin{align}
c^{\ell}_i = \sum^m_{j=1} a^{\ell}_{ij}(z^u_j+e_j). \label{eq:decinput}
\end{align}

The attention mechanism described here performs multiple attention
``hops'' per time step and considers which words have been previously
attended to. It is therefore different from single-step attention in
recurrent neural networks \cite{bahdanau-arxiv14}, where the attention
and weighted sum are computed over $\mathbf{z^u}$ only.

Our network uses multiple linear layers to project between the
embedding size $(f+f')$ and the convolution output size $2d$. They are
applied to~$\mathbf{e}$ before feeding it to the encoder, to the final
encoder output $\mathbf{z^u}$, to all decoder layers~$\mathbf{h^{\ell}}$
for the attention score computation, and to the final decoder output
$\mathbf{h^L}$ before the softmax.  We pad the input with~$k-1$ zero
vectors on both left and right side to ensure that the output of the
convolutional layers matches the input length. During decoding, we
ensure that the decoder does not have access to future information; we
start with $k$~zero vectors and shift the covolutional block to the
right after every prediction. The final decoder output $\mathbf{h^L}$
is used to compute the distribution over the target vocabulary $T$ as:
\begin{align}
  p(y_{i+1}|y_1,\ldots,y_i,D,t_D,\mathbf{t'}) = & \\
& \hspace*{-2cm}\mbox{softmax}(W_oh^L_i+b_o) \in \mathbb{R}^T. \label{eq:prediction}
\end{align}
We use layer normalization and weight initialization to
stabilize learning.

Our topic-enhanced model calibrates long-range dependencies with
globally salient content. As a result, it provides a better
alternative to vanilla convolutional sequence models
\cite{convseq2seq} and RNN-based summarization models \cite{see-acl17}
for capturing cross-document inferences and paraphrasing.  At the same
time it retains the computational advantages of convolutional
models. Each convolution block operates over a fixed-size window of
the input sequence, allowing for simultaneous encoding of the input,
ease in learning due to the fixed number of non-linearities and
transformations for words in the input sequence.



%% file: setup-evaluations.tex
\section{Experimental Setup}
\label{sec:setup}

In this section we present our experimental setup for assessing the
performance of our \textbf{T}opic-aware \textbf{Conv}olutional
\textbf{S}equence \textbf{to} \textbf{S}equence model which we call
\textsc{T-ConvS2S} for short.  We discuss implementation details and
present the systems used for comparison with our approach.

\paragraph{Comparison Systems} 
We report results with various systems which were all trained on the
XSum dataset to generate a one-line summary given an input news
article.  We compared \mbox{\textsc{T-ConvS2S}} against three
extractive systems: a baseline which randomly selects a sentence from
the input document (\textsc{random}), a baseline which simply selects
the leading sentence from the document (\textsc{lead}), and an oracle
which selects a single-best sentence in each document
(\textsc{ext-oracle}). The latter is often used as an upper bound for
extractive methods. We also compared our model against the RNN-based
abstractive systems introduced by \newcite{see-acl17}. In particular,
we experimented with an attention-based sequence to sequence model
(\textsc{Seq2Seq}), a pointer-generator model which allows to copy
words from the source text (\textsc{PtGen}), and a pointer-generator
model with a coverage mechanism to keep track of words that have been
summarized (\textsc{PtGen-Covg}). Finally, we compared our model
against the vanilla convolution sequence to sequence model
(\textsc{ConvS2S}) of \newcite{convseq2seq}.

\paragraph{Model Parameters and Optimization} 

We did not anonymize entities but worked on a lowercased version of
the XSum dataset. During training and at test time the input document
was truncated to 400~tokens and  the length of the summary limited to
90~tokens.

The LDA model \cite{Blei:2003:LDA} was trained on XSum documents
(training portion). We therefore obtained for each word a probability
distribution over topics which we used to estimate $\mathbf{t'}$; the
topic distribution $t_D$ can be inferred for any new document, at
training and test time. We explored several LDA configurations on
held-out data, and obtained best results with~512
topics. Table~\ref{fig:lda-topics} shows some of the topics learned by
the LDA model.


For \textsc{Seq2Seq}, \textsc{PtGen} and \textsc{PtGen-Covg}, we used
the best settings reported on the CNN and DailyMail data
\cite{see-acl17}.\footnote{We used the code available at
  \url{https://github.com/abisee/pointer-generator}.}  All three
models had 256~dimensional hidden states and 128~dimensional word
embeddings. They were trained using Adagrad \cite{Duchi:2011} with
learning rate~0.15 and an initial accumulator value of~0.1. We used
gradient clipping with a maximum gradient norm of~2, but did not use
any form of regularization.  We used the loss on the validation set to
implement early stopping. 

\begin{table}[t!]
  \center{\fontsize{10}{12}\selectfont 
    \begin{tabular}{@{}l@{\hspace{1.9ex}}p{6.7cm} }
      \thickhline 
      
      {T1:}& charge, court, murder, police, arrest, guilty, sentence, boy, bail, space, crown, trial \\
      
      {T2:}& church, abuse, bishop, child, catholic, gay, pope, school, christian, priest, cardinal \\
      
      {T3:}& council, people, government, local, housing, home, house, property, city, plan, authority \\
      
      {T4:}& clinton, party, trump, climate, poll, vote, plaid, election, debate, change, candidate, campaign\\
      
      {T5:}& country, growth, report, business, export, fall, bank, security, economy, rise, global, inflation\\
      
      {T6:}& hospital, patient, trust, nhs, people, care, health, service, staff, report, review, system, child\\

      \thickhline
    \end{tabular}     
  }
  \caption{Example topics learned by the LDA model on XSum documents
(training portion).}\label{fig:lda-topics}
\end{table}

For \textsc{ConvS2S}\footnote{We used the code available at
  \url{https://github.com/facebookresearch/fairseq-py}.} and
\textsc{T-ConvS2S}, we used 512 dimensional hidden states and 512
dimensional word and position embeddings. We trained our convolutional
models with Nesterov's accelerated gradient method
\cite{Sutskever:2013} using a momentum value of 0.99 and renormalized
gradients if their norm exceeded~0.1 \cite{Pascanu:2013}. We used a
learning rate of 0.10 and once the validation perplexity stopped
improving, we reduced the learning rate by an order of magnitude after
each epoch until it fell below~$10^{-4}$. We also applied a dropout
of~0.2 to the embeddings, the decoder outputs and the input of the
convolutional blocks. Gradients were normalized by the number of
non-padding tokens per mini-batch. We also used weight normalization
for all layers except for lookup tables.

All neural models, including ours and those based on RNNs \cite{see-acl17}
had a vocabulary of 50,000 words and were trained on a single Nvidia
M40 GPU with a batch size of~32 sentences.  Summaries at test time
were obtained using beam search (with beam size 10).

\section{Results}
\label{sec:results}

\paragraph{Automatic Evaluation} 

We report results using automatic metrics in Table~\ref{tab:rouge}.
We evaluated summarization quality using F$_1$ $\mbox{ROUGE}$
\cite{rouge}. Unigram and bigram overlap ($\mbox{ROUGE-1}$ and
$\mbox{ROUGE-2}$) are a proxy for assessing informativeness and the
longest common subsequence ($\mbox{ROUGE-L}$) represents
fluency.\footnote{We used \texttt{pyrouge} to compute all ROUGE
  scores, with parameters ``-a -c 95 -m -n 4 -w 1.2.''}

On the XSum dataset, \textsc{Seq2Seq} outperforms the \textsc{lead}
and \textsc{random} baselines by a large margin. \textsc{PtGen}, a
\textsc{Seq2Seq} model with a ``copying'' mechanism outperforms
\textsc{ext-oracle}, a ``perfect'' extractive system on ROUGE-2 and
ROUGE-L. This is in sharp contrast to the performance of these models
on CNN/DailyMail \cite{see-acl17} and  Newsroom datasets
\cite{newsroom-naacl18}, where they fail to outperform the
\textsc{lead}. The result provides further evidence that XSum is a
good testbed for abstractive summarization. \textsc{PtGen-Covg}, the
best performing abstractive system on the CNN/DailyMail datasets, does
not do well.  We believe that the coverage mechanism is more useful
when generating multi-line summaries and is basically redundant for
extreme summarization. 

\begin{table}[t]
  \begin{center}{\fontsize{8.5}{11}\selectfont 
  \begin{tabular}{ l | c c c }
    \thickhline
    Models & R1 & R2 & RL \\ \thickhline
    Random & 15.16 & 1.78 & 11.27 \\ 
    \textsc{lead} & 16.30 & 1.60 & 11.95 \\ 
    \textsc{ext-oracle} & 29.79 & 8.81 & 22.66 \\ \hline
    \textsc{Seq2Seq} &  28.42 & 8.77 &  22.48 \\
    \textsc{PtGen} & 29.70 & 9.21 & 23.24 \\ 
    \textsc{PtGen-Covg} & 28.10 & 8.02 & 21.72 \\ 
    \hline
    \textsc{ConvS2S} & 31.27 & 11.07 & 25.23 \\  
    \textsc{T-ConvS2S} (enc$_{t'}$) & 31.71 & 11.38 & 25.56 \\
    \textsc{T-ConvS2S} (enc$_{t'}$, dec$_{t_D}$) & 31.71 & 11.34 & 25.61 \\
    \textsc{T-ConvS2S} (enc$_{(t',t_D)}$) & 31.61 & 11.30 & 25.51 \\
    \textsc{T-ConvS2S} (enc$_{(t',t_D)}$, dec$_{t_D}$)  & \textbf{31.89} & \textbf{11.54} & \textbf{25.75} \\ \thickhline 
    
  \end{tabular}
  }\end{center}
\caption{ROUGE results on  XSum test set. We report \mbox{ROUGE-1} (R1), ROUGE-2
  (R2), and ROUGE-L (RL) F$_1$  scores. Extractive systems are in the
  upper block, RNN-based   abstractive systems are in the middle block,
  and convolutional    abstractive systems are in the bottom
  block. \label{tab:rouge}} 
\end{table}

\begin{table}
  \begin{center}{\fontsize{8.5}{11}\selectfont 
  \begin{tabular}{ l | c c c c } 
    \thickhline
    \multirow{2}{*}{Models} & \multicolumn{4}{c}{\% of novel n-grams in generated summaries}  \\
    & unigrams & bigrams & trigrams & 4-grams  \\ \thickhline 
    \textsc{lead} & 0.00 & 0.00 & 0.00 & 0.00 \\
    \textsc{ext-oracle} & 0.00 & 0.00 & 0.00 & 0.00 \\
    \textsc{PtGen} & 27.40 & 73.33 & 90.43 & 96.04 \\ 
    \textsc{ConvS2S} & \textbf{31.26} & \textbf{79.50} & \textbf{94.28} & \textbf{98.10}\\
    \textsc{T-ConvS2S}  & 30.73 & 79.18 & 94.10 & 98.03 \\ \hline
    \textsc{gold} & 35.76 & 83.45  & 95.50  & 98.49 \\ \thickhline
  \end{tabular}}
  \end{center}
  \caption{Proportion of novel $n$-grams in summaries generated by various models on the XSum test set.
    \label{tab:novelngram-xsum-models}}
\end{table}

\begin{table*}[t!]
  \center{ \footnotesize
    \begin{tabular}{l p{10.5cm} l}
      \thickhline 
      
      \textsc{ext-oracle} &Caroline Pidgeon is the Lib Dem candidate, Sian Berry will contest the election for the Greens and UKIP has chosen its culture spokesman Peter Whittle. &[34.1, 20.5, 34.1]\\
            
      {\textsc{PtGen}} & UKIP leader \textcolor{red}{Nigel Goldsmith} has been \textcolor{red}{elected} as the \textcolor{red}{new mayor of London} to elect a new conservative MP. &[45.7, 6.1, 28.6]  \\
      
      {\textsc{ConvS2S}}&  London mayoral candidate \textcolor{midnightblue}{Zac Goldsmith} has been \textcolor{red}{elected} as the \textcolor{red}{new mayor of London}.& [53.3, 21.4, 26.7]  \\

      {\textsc{T-ConvS2S}}& Former London mayoral candidate \textcolor{midnightblue}{Zac Goldsmith} has been \textcolor{midnightblue}{chosen to stand} in \textcolor{midnightblue}{the London mayoral election}. &[50.0, 26.7, 37.5] \\
      
      {\textsc{gold}}& \textcolor{midnightblue}{Zac Goldsmith} will \textcolor{midnightblue}{contest} the 2016 \textcolor{midnightblue}{London mayoral election} for the conservatives, it has been announced. \\
      
      {Questions}& (1) {Who will \textcolor{midnightblue}{contest} for
        the conservatives?} (\textcolor{midnightblue}{Zac
        Goldsmith})\\
 &  (2) {For what election will he/she contest?} (\textcolor{midnightblue}{The London mayoral election}) \\
            
      \hline
      {\textsc{ext-oracle}} &North-east rivals Newcastle are the only team below them in the Premier League table. &[35.3, 18.8, 35.3] \\
            
      {\textsc{PtGen}} &Sunderland have \textcolor{red}{appointed} former \textcolor{midnightblue}{Sunderland boss} \textcolor{midnightblue}{Dick Advocaat} as manager at the end of the season to sign a new deal. &[45.0, 10.5, 30.0] \\
      
      {\textsc{ConvS2S}} &Sunderland have \textcolor{red}{sacked} \textcolor{midnightblue}{manager} \textcolor{midnightblue}{Dick Advocaat} after less than three months in charge. &[25.0, 6.7, 18.8] \\

      {\textsc{T-ConvS2S}}& \textcolor{midnightblue}{Dick Advocaat} has \textcolor{midnightblue}{resigned} as \textcolor{midnightblue}{Sunderland manager} until the end of the season.& [56.3, 33.3, 56.3] \\
      
      {\textsc{gold}} &\textcolor{midnightblue}{Dick Advocaat} has \textcolor{midnightblue}{resigned} as \textcolor{midnightblue}{Sunderland boss}, with the team yet to win in the Premier League this season. \\
      
      Questions& (1) {Who has \textcolor{midnightblue}{resigned}?}
      (\textcolor{midnightblue}{Dick Advocaat}) \\& (2) {From what post has he/she resigned?} (\textcolor{midnightblue}{Sunderland boss}) \\
            
      \hline
      
      {\textsc{ext-oracle}}& The Greater Ardoyne residents collective (GARC) is protesting against an agreement aimed at resolving a long-running dispute in the area. &[26.7, 9.3, 22.2]  \\

      {\textsc{PtGen}} &A residents' group has been granted permission for GARC to hold a parade on \textcolor{red}{the outskirts of Crumlin, County Antrim}. &[28.6, 5.0, 28.6] \\

      {\textsc{ConvS2S}} &A protest has been held in the \textcolor{midnightblue}{Republic of Ireland} calling for \textcolor{red}{an end to parading} parading in \textcolor{midnightblue}{North Belfast}. &[42.9, 20.0, 33.3] \\

      {\textsc{T-ConvS2S}}& A protest has been held in \textcolor{midnightblue}{North Belfast} over a protest against \textcolor{midnightblue}{the Orange Order} in \textcolor{midnightblue}{North Belfast}.& [45.0, 26.3, 45.0] \\

      {\textsc{gold}} &Church leaders have appealed to a nationalist residents' group to call off a protest against \textcolor{midnightblue}{an Orange Order parade} in \textcolor{midnightblue}{North Belfast}. \\

      Questions &(1) {Where is the protest supposed to happen?}
      (\textcolor{midnightblue}{North Belfast}) \\ 
&(2) {What are they protesting against?} (\textcolor{midnightblue}{An Orange Order parade}) \\
      \thickhline

    \end{tabular}     
  }
  \caption{Example output summaries on the XSum test set with [ROUGE-1,
    ROUGE-2 and ROUGE-L] scores, goldstandard reference, and
    corresponding questions. Words highlighted in blue are either the
    right answer or constitute appropriate context for inferring it;
    words in red lead to the wrong
    answer. }\label{tab:erroranalysis}
\end{table*}

\textsc{ConvS2S}, the convolutional variant of \textsc{Seq2Seq},
significantly outperforms all \mbox{RNN-based} abstractive systems. We
hypothesize that its superior performance stems from the ability to
better represent document content (i.e.,~by capturing long-range
dependencies).
Table~\ref{tab:rouge} shows several variants of \textsc{T-ConvS2S}
including an encoder network enriched with information about how
topical a word is on its own (enc$_{t'}$) or in the document
(enc$_{(t',t_D)}$).  We also experimented with various decoders by
conditioning every prediction on the topic of the document, basically
encouraging the summary to be in the same theme as the document
(dec$_{t_D}$) or letting the decoder decide the theme of the
summary. Interestingly, all four \textsc{T-ConvS2S} variants
outperform \textsc{ConvS2S}. \textsc{T-ConvS2S} performs best when
both encoder and decoder are constrained by the document topic
(enc$_{(t',t_D)}$,dec$_{t_D}$). In the remainder of the paper, we
refer to this variant as \textsc{T-ConvS2S}.

We further assessed the extent to which various models are able to
perform rewriting by generating genuinely abstractive
summaries. Table~\ref{tab:novelngram-xsum-models} shows the proportion
of novel $n$-grams for \textsc{lead}, \textsc{ext-oracle},
\textsc{PtGen}, \textsc{ConvS2S}, and \textsc{T-ConvS2S}.  As can be
seen, the convolutional models exhibit the highest proportion of novel
$n$-grams. We should also point out that the summaries being evaluated
have on average comparable lengths; the summaries generated by
\textsc{PtGen} contain 22.57 words, those generated by
\textsc{ConvS2S} and \textsc{T-ConvS2S} have 20.07 and 20.22 words,
respectively, while \textsc{gold} summaries are the longest with~23.26
words.  Interestingly, \textsc{PtGen} trained on XSum only copies 4\%
of 4-grams in the source document, 10\% of trigrams, 27\% of bigrams,
and 73\% of unigrams. This is in sharp contrast to \textsc{PtGen}
trained on CNN/DailyMail exhibiting mostly extractive patterns; it
copies more than 85\% of 4-grams in the source document, 90\% of
trigrams, 95\% of bigrams, and 99\% of unigrams \cite{see-acl17}. This
result further strengthens our hypothesis that XSum is a good testbed
for abstractive methods.


\paragraph{Human Evaluation}
In addition to automatic evaluation using ROUGE which can be
misleading when used as the only means to assess the informativeness
of summaries \cite{schluter:2017:EACLshort}, we also evaluated system
output by eliciting human judgments in two ways. 

In our first experiment, participants were asked to compare summaries
produced from the \textsc{ext-oracle} baseline, \textsc{PtGen}, the
best performing system of \citet{see-acl17}, \textsc{ConvS2S}, our
topic-aware model \textsc{T-ConvS2S}, and the human-authored gold
summary (\textsc{gold}). We did not include extracts from the
\textsc{lead} as they were significantly inferior to other models.

The study was conducted on the Amazon Mechanical Turk platform using
\textit{Best-Worst Scaling} (BWS; \citealt{louviere1991best,
  louviere2015best}), a less labor-intensive alternative to paired
comparisons that has been shown to produce more reliable results than
rating scales \cite{bestworstscaling}. Participants were presented
with a document and summaries generated from two out of five systems
and were asked to decide which summary was \textit{better} and which one
was \textit{worse} in order of informativeness (does the summary
capture important information in the document?) and fluency (is the
summary written in well-formed English?). Examples of system summaries
are shown in Table~\ref{tab:erroranalysis}.  We randomly selected
50~documents from the XSum test set and compared all possible
combinations of two out of five systems for each document. We
collected judgments from three different participants for each
comparison. The order of summaries was randomized per document and the
order of documents per participant.

\begin{table}[t]
  \center{\small
    \begin{tabular}{lrc}
      \thickhline
      Models  & Score & QA\\ \hline
      \textsc{ext-oracle} &  -0.121 & 15.70\\
      \textsc{PtGen} & -0.218& 21.40 \\
      \textsc{ConvS2S}  &
      -0.130 & 30.90 \\
      \textsc{T-ConvS2S} & \textbf{0.037} & \textbf{46.05}\\ \hline
      \textsc{gold} & 0.431 & 97.23\\
      \thickhline
    \end{tabular}}
  \caption{System ranking according to human judgments  and QA-based
    evaluation.\label{tab:heval-pref}}
\end{table}

The score of a system was computed as the percentage of times it was
chosen as best minus the percentage of times it was selected as
worst. The scores range from -1 (worst) to 1 (best) and are shown in
Table~\ref{tab:heval-pref}. Perhaps unsurprisingly human-authored
summaries were considered best, whereas, \textsc{T-ConvS2S} was ranked
2nd followed by \textsc{ext-oracle} and
\textsc{ConvS2S}. \textsc{PtGen} was ranked worst with the lowest
score of $-0.218$. We carried out pairwise comparisons between all
models to assess whether system differences are statistically
significant. \textsc{gold} is significantly different from all other
systems and \textsc{T-ConvS2S} is significantly different from
\textsc{ConvS2S} and \textsc{PtGen} (using a one-way ANOVA with
posthoc Tukey HSD tests; $p < 0.01$). All other differences are not
statistically significant.


%

For our second experiment we used a question-answering (QA) paradigm
\cite{Clarke:Lapata:2010,narayan-rank18} to assess the degree to which
the models retain key information from the document. We used the same
50 documents as in our first elicitation study.  We wrote two
fact-based questions per document, just by reading the summary, under
the assumption that it highlights the most important content of the
news article. Questions were formulated so as not to reveal answers to
subsequent questions.  We created 100~questions in total (see
Table~\ref{tab:erroranalysis} for examples). Participants read the
output summaries and answered the questions as best they could without
access to the document or the gold summary. The more questions can be
answered, the better the corresponding system is at summarizing the
document as a whole.  Five participants answered questions for each
summary.

We followed the scoring mechanism introduced in
\newcite{Clarke:Lapata:2010}. A correct answer was marked with a score
of one, partially correct answers with a score of~0.5, and zero
otherwise. The final score for a system is the average of all its
question scores. Answers again were elicited using Amazon's Mechanical
Turk crowdsourcing platform. We uploaded the data in batches (one
system at a time) to ensure that the same participant does not
evaluate summaries from different systems on the same set of
questions.





Table~\ref{tab:heval-pref} shows the results of the QA
evaluation. Based on summaries generated by \mbox{\textsc{T-ConvS2S}},
participants can answer $46.05\%$ of the questions
correctly. Summaries generated by \textsc{ConvS2S}, \textsc{PtGen} and
\textsc{ext-oracle} provide answers to $30.90\%$, $21.40\%$, and
$15.70\%$ of the questions, respectively. Pairwise differences between
systems are all statistically significant ($p < 0.01$) with the
exception of \textsc{PtGen} and \textsc{ext-oracle}.
\textsc{ext-oracle} performs poorly on both QA and rating
evaluations. The examples in Table~\ref{tab:erroranalysis} indicate
that \textsc{ext-oracle} is often misled by selecting a sentence with
the highest ROUGE (against the gold summary), but ROUGE itself does
not ensure that the summary retains the most important information
from the document.  The QA evaluation further emphasizes that in order
for the summary to be felicitous, information needs to be embedded in
the appropriate context. For example, \textsc{ConvS2S} and
\textsc{PtGen} will fail to answer the question ``Who has resigned?''
(see Table~\ref{tab:erroranalysis} second block) despite containing
the correct answer ``Dick Advocaat'' due to the wrong
context. \mbox{\textsc{T-ConvS2S}} is able to extract
important entities from the document with the right theme.

%% file: extreme-sum.bbl
\begin{thebibliography}{42}
\expandafter\ifx\csname natexlab\endcsname\relax\def\natexlab#1{#1}\fi

\bibitem[{Bahdanau et~al.(2015)Bahdanau, Cho, and Bengio}]{bahdanau-arxiv14}
Dzmitry Bahdanau, Kyunghyun Cho, and Yoshua Bengio. 2015.
\newblock Neural machine translation by jointly learning to align and
  translate.
\newblock In \emph{Proceedings of the 3rd International Conference on Learning
  Representations}, San Diego, California, USA.

\bibitem[{Blei et~al.(2003)Blei, Ng, and Jordan}]{Blei:2003:LDA}
David~M. Blei, Andrew~Y. Ng, and Michael~I. Jordan. 2003.
\newblock Latent dirichlet allocation.
\newblock \emph{The Journal of Machine Learning Research}, 3:993--1022.

\bibitem[{Celikyilmaz et~al.(2018)Celikyilmaz, Bosselut, He, and
  Choi}]{asli-multiagent18}
Asli Celikyilmaz, Antoine Bosselut, Xiaodong He, and Yejin Choi. 2018.
\newblock Deep communicating agents for abstractive summarization.
\newblock In \emph{Proceedings of the 16th Annual Conference of the North
  American Chapter of the Association for Computational Linguistics: Human
  Language Technologies}, New Orleans, USA.

\bibitem[{Chen et~al.(2016)Chen, Zhu, Ling, Wei, and Jiang}]{chenIjcai-16}
Qian Chen, Xiaodan Zhu, Zhenhua Ling, Si~Wei, and Hui Jiang. 2016.
\newblock Distraction-based neural networks for modeling documents.
\newblock In \emph{Proceedings of the 25th International Joint Conference on
  Artificial Intelligence}, pages 2754--2760, New York, USA.

\bibitem[{Cheng and Lapata(2016)}]{jp-acl16}
Jianpeng Cheng and Mirella Lapata. 2016.
\newblock Neural summarization by extracting sentences and words.
\newblock In \emph{Proceedings of the 54th Annual Meeting of the Association
  for Computational Linguistics}, pages 484--494, Berlin, Germany.

\bibitem[{Clarke and Lapata(2010)}]{Clarke:Lapata:2010}
James Clarke and Mirella Lapata. 2010.
\newblock Discourse constraints for document compression.
\newblock \emph{Computational Linguistics}, 36(3):411--441.

\bibitem[{Dauphin et~al.(2017)Dauphin, Fan, Auli, and
  Grangier}]{pmlr-v70-dauphin17a}
Yann~N. Dauphin, Angela Fan, Michael Auli, and David Grangier. 2017.
\newblock Language modeling with gated convolutional networks.
\newblock In \emph{Proceedings of the 34th International Conference on Machine
  Learning}, pages 933--941, Sydney, Australia.

\bibitem[{Dieng et~al.(2017)Dieng, Wang, Gao, and Paisley}]{dieng-iclr17}
Adji~B. Dieng, Chong Wang, Jianfeng Gao, and John Paisley. 2017.
\newblock Topicrnn: A recurrent neural network with long-range semantic
  dependency.
\newblock In \emph{Proceedings of the 5th International Conference on Learning
  Representations}, Toulon, France.

\bibitem[{Duchi et~al.(2011)Duchi, Hazan, and Singer}]{Duchi:2011}
John Duchi, Elad Hazan, and Yoram Singer. 2011.
\newblock Adaptive subgradient methods for online learning and stochastic
  optimization.
\newblock \emph{Journal of Machine Learning Research}, 12:2121--2159.

\bibitem[{Durrett et~al.(2016)Durrett, Berg-Kirkpatrick, and
  Klein}]{durrett-nyt-ext}
Greg Durrett, Taylor Berg-Kirkpatrick, and Dan Klein. 2016.
\newblock Learning-based single-document summarization with compression and
  anaphoricity constraints.
\newblock In \emph{Proceedings of the 54th Annual Meeting of the Association
  for Computational Linguistics}, pages 1998--2008, Berlin, Germany.

\bibitem[{Fan et~al.(2017)Fan, Grangier, and Auli}]{Fan2017Controllable}
Angela Fan, David Grangier, and Michael Auli. 2017.
\newblock Controllable abstractive summarization.
\newblock \emph{CoRR}, abs/1711.05217.

\bibitem[{Fan et~al.(2018)Fan, Lewis, and Dauphin}]{fan-hier-gen}
Angela Fan, Mike Lewis, and Yann Dauphin. 2018.
\newblock Hierarchical neural story generation.
\newblock In \emph{Proceedings of the 56th Annual Meeting of the Association
  for Computational Linguistics}, Melbourne, Australia.

\bibitem[{Gehring et~al.(2017{\natexlab{a}})Gehring, Auli, Grangier, and
  Dauphin}]{convenc_mt}
Jonas Gehring, Michael Auli, David Grangier, and Yann Dauphin.
  2017{\natexlab{a}}.
\newblock A convolutional encoder model for neural machine translation.
\newblock In \emph{Proceedings of the 55th Annual Meeting of the Association
  for Computational Linguistics}, pages 123--135, Vancouver, Canada.

\bibitem[{Gehring et~al.(2017{\natexlab{b}})Gehring, Auli, Grangier, Yarats,
  and Dauphin}]{convseq2seq}
Jonas Gehring, Michael Auli, David Grangier, Denis Yarats, and Yann~N. Dauphin.
  2017{\natexlab{b}}.
\newblock Convolutional sequence to sequence learning.
\newblock In \emph{Proceedings of the 34th International Conference on Machine
  Learning}, volume~70, pages 1243--1252, Sydney, Australia.

\bibitem[{Ghosh et~al.(2016)Ghosh, Vinyals, Strope, Roy, Dean, and
  Heck}]{Ghosh2016ContextualL}
Shalini Ghosh, Oriol Vinyals, Brian Strope, Scott Roy, Tom Dean, and Larry
  Heck. 2016.
\newblock Contextual {LSTM (CLSTM)} models for large scale {NLP} tasks.
\newblock \emph{CoRR}, abs/1602.06291.

\bibitem[{Grusky et~al.(2018)Grusky, Naaman, and Artzi}]{newsroom-naacl18}
Max Grusky, Mor Naaman, and Yoav Artzi. 2018.
\newblock {NEWSROOM: A} dataset of 1.3 million summaries with diverse
  extractive strategies.
\newblock In \emph{Proceedings of the 16th Annual Conference of the North
  American Chapter of the Association for Computational Linguistics: Human
  Language Technologies}, New Orleans, USA.

\bibitem[{He et~al.(2016)He, Zhang, Ren, and Sun}]{He2016DeepRL}
Kaiming He, Xiangyu Zhang, Shaoqing Ren, and Jian Sun. 2016.
\newblock Deep residual learning for image recognition.
\newblock In \emph{Proceedings of the IEEE Conference on Computer Vision and
  Pattern Recognition}, pages 770--778, Las Vegas, USA.

\bibitem[{Hermann et~al.(2015)Hermann, Ko\v{c}isk\'{y}, Grefenstette, Espeholt,
  Kay, Suleyman, and Blunsom}]{hermann-nips15}
Karl~Moritz Hermann, Tom\'{a}\v{s} Ko\v{c}isk\'{y}, Edward Grefenstette, Lasse
  Espeholt, Will Kay, Mustafa Suleyman, and Phil Blunsom. 2015.
\newblock Teaching machines to read and comprehend.
\newblock In \emph{Advances in Neural Information Processing Systems 28}, pages
  1693--1701. Morgan, Kaufmann.

\bibitem[{Kiritchenko and Mohammad(2017)}]{bestworstscaling}
Svetlana Kiritchenko and Saif Mohammad. 2017.
\newblock Best-worst scaling more reliable than rating scales: A case study on
  sentiment intensity annotation.
\newblock In \emph{Proceedings of the 55th Annual Meeting of the Association
  for Computational Linguistics}, pages 465--470, Vancouver, Canada.

\bibitem[{Lin and Hovy(2003)}]{rouge}
Chin-Yew Lin and Eduard Hovy. 2003.
\newblock Automatic evaluation of summaries using n-gram co-occurrence
  statistics.
\newblock In \emph{Proceedings of the 2003 Human Language Technology Conference
  of the North American Chapter of the Association for Computational
  Linguistics}, pages 71--78, Edmonton, Canada.

\bibitem[{Louviere et~al.(2015)Louviere, Flynn, and Marley}]{louviere2015best}
Jordan~J Louviere, Terry~N Flynn, and Anthony Alfred~John Marley. 2015.
\newblock \emph{Best-worst scaling: Theory, methods and applications}.
\newblock Cambridge University Press.

\bibitem[{Louviere and Woodworth(1991)}]{louviere1991best}
Jordan~J Louviere and George~G Woodworth. 1991.
\newblock Best-worst scaling: A model for the largest difference judgments.
\newblock \emph{University of Alberta: Working Paper}.

\bibitem[{Mani(2001)}]{mani2001automatic}
Inderjeet Mani. 2001.
\newblock \emph{Automatic Summarization}.
\newblock Natural language processing. John Benjamins Publishing Company.

\bibitem[{Mikolov and Zweig(2012)}]{mikolovZ12}
Tomas Mikolov and Geoffrey Zweig. 2012.
\newblock Context dependent recurrent neural network language model.
\newblock In \emph{Proceedings of the Spoken Language Technology Workshop},
  pages 234--239. IEEE.

\bibitem[{Nallapati et~al.(2017)Nallapati, Zhai, and Zhou}]{nallapati17}
Ramesh Nallapati, Feifei Zhai, and Bowen Zhou. 2017.
\newblock {SummaRuNNer}: {A} recurrent neural network based sequence model for
  extractive summarization of documents.
\newblock In \emph{Proceedings of the 31st AAAI Conference on Artificial
  Intelligence}, pages 3075--3081, San Francisco, California USA.

\bibitem[{Nallapati et~al.(2016)Nallapati, Zhou, dos Santos,
  G{\"{u}}l{\c{c}}ehre, and Xiang}]{nallapati-signll16}
Ramesh Nallapati, Bowen Zhou, C{\'{\i}}cero~Nogueira dos Santos, {\c{C}}aglar
  G{\"{u}}l{\c{c}}ehre, and Bing Xiang. 2016.
\newblock Abstractive text summarization using sequence-to-sequence {RNNs} and
  beyond.
\newblock In \emph{Proceedings of the 20th SIGNLL Conference on Computational
  Natural Language Learning}, pages 280--290, Berlin, Germany.

\bibitem[{Narayan et~al.(2018{\natexlab{a}})Narayan, Cardenas,
  Papasarantopoulos, Cohen, Lapata, Yu, and Chang}]{narayan-sidenet18}
Shashi Narayan, Ronald Cardenas, Nikos Papasarantopoulos, Shay~B. Cohen,
  Mirella Lapata, Jiangsheng Yu, and Yi~Chang. 2018{\natexlab{a}}.
\newblock Document modeling with external attention for sentence extraction.
\newblock In \emph{Proceedings of the 56th Annual Meeting of the Association
  for Computational Linguistics}, Melbourne, Australia.

\bibitem[{Narayan et~al.(2018{\natexlab{b}})Narayan, Cohen, and
  Lapata}]{narayan-rank18}
Shashi Narayan, Shay~B. Cohen, and Mirella Lapata. 2018{\natexlab{b}}.
\newblock Ranking sentences for extractive summarization with reinforcement
  learning.
\newblock In \emph{Proceedings of the 16th Annual Conference of the North
  American Chapter of the Association for Computational Linguistics: Human
  Language Technologies}, New Orleans, USA.

\bibitem[{Narayan et~al.(2017)Narayan, Papasarantopoulos, Cohen, and
  Lapata}]{narayan-arxiv17}
Shashi Narayan, Nikos Papasarantopoulos, Shay~B. Cohen, and Mirella Lapata.
  2017.
\newblock Neural extractive summarization with side information.
\newblock \emph{CoRR}, abs/1704.04530.

\bibitem[{Nenkova(2005)}]{nenkova-05}
Ani Nenkova. 2005.
\newblock Automatic text summarization of newswire: Lessons learned from the
  {D}ocument {U}nderstanding {C}onference.
\newblock In \emph{Proceedings of the 29th National Conference on Artificial
  Intelligence}, pages 1436--1441, Pittsburgh, Pennsylvania, USA.

\bibitem[{Nenkova and McKeown(2011)}]{Nenkova:McKeown:2011}
Ani Nenkova and Kathleen McKeown. 2011.
\newblock Automatic summarization.
\newblock \emph{Foundations and Trends in Information Retrieval},
  5(2--3):103--233.

\bibitem[{Pascanu et~al.(2013)Pascanu, Mikolov, and Bengio}]{Pascanu:2013}
Razvan Pascanu, Tomas Mikolov, and Yoshua Bengio. 2013.
\newblock On the difficulty of training recurrent neural networks.
\newblock In \emph{Proceedings of the 30th International Conference on
  International Conference on Machine Learning}, pages 1310--1318, Atlanta, GA,
  USA.

\bibitem[{Pasunuru and Bansal(2018)}]{Pasunuru-multireward18}
Ramakanth Pasunuru and Mohit Bansal. 2018.
\newblock Multi-reward reinforced summarization with saliency and entailment.
\newblock In \emph{Proceedings of the 16th Annual Conference of the North
  American Chapter of the Association for Computational Linguistics: Human
  Language Technologies}, New Orleans, USA.

\bibitem[{Paulus et~al.(2018)Paulus, Xiong, and Socher}]{paulus-socher-arxiv17}
Romain Paulus, Caiming Xiong, and Richard Socher. 2018.
\newblock A deep reinforced model for abstractive summarization.
\newblock In \emph{Proceedings of the 6th International Conference on Learning
  Representations}, Vancouver, BC, Canada.

\bibitem[{Rush et~al.(2015)Rush, Chopra, and Weston}]{rush-acl15}
Alexander~M. Rush, Sumit Chopra, and Jason Weston. 2015.
\newblock A neural attention model for abstractive sentence summarization.
\newblock In \emph{Proceedings of the 2015 Conference on Empirical Methods in
  Natural Language Processing}, pages 379--389, Lisbon, Portugal.

\bibitem[{Sandhaus(2008)}]{nytcorpus}
Evan Sandhaus. 2008.
\newblock {The New York Times Annotated Corpus}.
\newblock \emph{Linguistic Data Consortium, Philadelphia}, 6(12).

\bibitem[{Schluter(2017)}]{schluter:2017:EACLshort}
Natalie Schluter. 2017.
\newblock The limits of automatic summarisation according to rouge.
\newblock In \emph{Proceedings of the 15th Conference of the European Chapter
  of the Association for Computational Linguistics: Short Papers}, pages
  41--45, Valencia, Spain.

\bibitem[{See et~al.(2017)See, Liu, and Manning}]{see-acl17}
Abigail See, Peter~J. Liu, and Christopher~D. Manning. 2017.
\newblock Get to the point: {S}ummarization with pointer-generator networks.
\newblock In \emph{Proceedings of the 55th Annual Meeting of the Association
  for Computational Linguistics}, pages 1073--1083, Vancouver, Canada.

\bibitem[{Shi et~al.(2016)Shi, Knight, and Yuret}]{D16-1248}
Xing Shi, Kevin Knight, and Deniz Yuret. 2016.
\newblock Why neural translations are the right length.
\newblock In \emph{Proceedings of the 2016 Conference on Empirical Methods in
  Natural Language Processing}, pages 2278--2282, Austin, Texas.

\bibitem[{Sukhbaatar et~al.(2015)Sukhbaatar, szlam, Weston, and
  Fergus}]{NIPS2015_5846}
Sainbayar Sukhbaatar, arthur szlam, Jason Weston, and Rob Fergus. 2015.
\newblock End-to-end memory networks.
\newblock In \emph{Advances in Neural Information Processing Systems 28}, pages
  2440--2448. Morgan, Kaufmann.

\bibitem[{Sutskever et~al.(2013)Sutskever, Martens, Dahl, and
  Hinton}]{Sutskever:2013}
Ilya Sutskever, James Martens, George Dahl, and Geoffrey Hinton. 2013.
\newblock On the importance of initialization and momentum in deep learning.
\newblock In \emph{Proceedings of the 30th International Conference on
  International Conference on Machine Learning}, pages 1139--1147, Atlanta, GA,
  USA.

\bibitem[{Tan and Wan(2017)}]{tanwan-acl17}
Jiwei Tan and Xiaojun Wan. 2017.
\newblock Abstractive document summarization with a graph-based attentional
  neural model.
\newblock In \emph{Proceedings of the 55th Annual Meeting of the Association
  for Computational Linguistics}, pages 1171--1181, Vancouver, Canada.

\end{thebibliography}
